\documentclass[11pt]{article}

\usepackage[final]{acl}
\usepackage{times}
\usepackage{latexsym}
\usepackage[T1]{fontenc}
\usepackage[utf8]{inputenc}
\usepackage{microtype}
\usepackage{inconsolata}
\usepackage{graphicx}
\usepackage{booktabs}
\usepackage{makecell}
\usepackage{multirow}
\usepackage{array}
\usepackage{xcolor}
\usepackage{tikz}
\usepackage{tcolorbox}
\usetikzlibrary{arrows.meta,positioning,fit}

\newcommand{\SR}{\textsc{Self-Recall}}
\newcommand{\QR}{\textsc{Question-Recall}}

\definecolor{cutoffblue}{RGB}{38,92,140}
\definecolor{recallgreen}{RGB}{63,128,99}
\definecolor{eventorange}{RGB}{186,103,35}
\definecolor{panelgray}{RGB}{245,247,250}

\tcbuselibrary{breakable,skins}
\newtcolorbox{promptbox}[1]{
  colback=gray!4,
  colframe=gray!55,
  title=#1,
  fonttitle=\normalsize\bfseries,
  fontupper=\small,
  boxrule=0.4pt,
  arc=2pt,
  left=4pt,
  right=4pt,
  top=3pt,
  bottom=3pt,
  toptitle=3pt,
  bottomtitle=3pt
}

\usepackage{array} 
\usepackage[normalem]{ulem} 
\newcommand{\thickhline}{\noalign{\hrule height 1.2pt}}
\newcommand{\thinhline}{\noalign{\hrule height 0.4pt}}

\title{Can LLMs Be Constrained to the Past?\\
Improving Knowledge Cutoff through Recall-Based Prompting}

\author{
    Michiro Asai, 
    Ailiang Lin, 
    Yu Kishimoto,
    Takao Obi,
    Satoshi Kosugi,\\
    \textbf{Kotaro Funakoshi},
    \textbf{Manabu Okumura}\\
    Institute of Science Tokyo\\
\texttt{\{asai,linailiang,ky,obi,kosugi,funakoshi,oku\}@lr.first.iir.isct.ac.jp}
}

\begin{document}
\maketitle

\begin{abstract}
Prompted knowledge cutoff instructs a large language model (LLM) to act as if information beyond a specified cutoff date were unavailable. However, prior work mainly relies on direct-answer generation, which struggles when post-cutoff knowledge is not explicitly queried but is only causally related to the question. To address this limitation, we propose two recall-based prompting strategies: Self-Recall (SR), which asks the model to restate its cutoff constraint, and Question-Recall (QR), which requires the model to recall question-relevant information valid under the cutoff. Across three existing benchmarks, our methods outperform both direct-answer prompting and conventional step-by-step reasoning baselines, with particularly strong improvements on counterfactual questions. To investigate robustness across different cutoff settings, we further construct the Multi-cutoff Historical Event Benchmark (MHEB), which evaluates the same question under multiple cutoff years. Results show that knowledge cutoff performance varies with cutoff distance, while combining SR and QR consistently yields the best performance.
\end{abstract}

\begin{figure*}[htbp]
\centering
\includegraphics[width=0.99\textwidth]{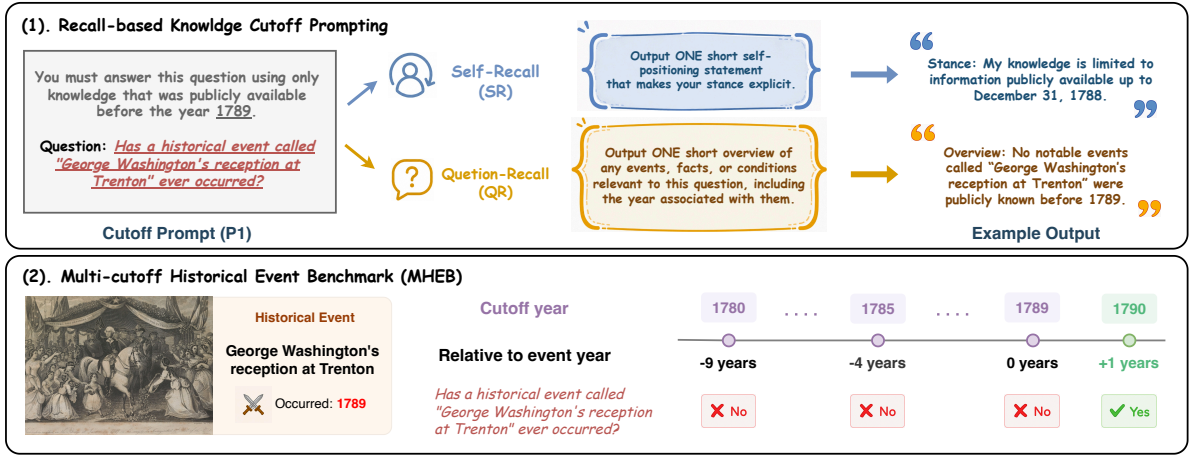}
\caption{(1) Proposed recall-based prompting strategies: Self-Recall (SR) and Question-Recall (QR), illustrated with \textit{cutoff year} $1789$. (2) The proposed Multi-cutoff Historical Event Benchmark (MHEB).}
\label{fig:overall}
\end{figure*}

\section{Introduction}
Large language models (LLMs) are increasingly applied to temporal prediction tasks such as stock price forecasting, climate modeling, and traffic prediction~\cite{yu2023harnessing,jin2024time}. However, LLMs are pretrained on large-scale corpora collected over long time spans, raising a fundamental concern of data contamination: evaluation on examples predating the model's actual knowledge cutoff may probe memorized knowledge rather than genuine temporal reasoning~\cite{roberts2023cutoff,dong-etal-2024-generalization}. Therefore, constraining an LLM to answer based only on knowledge available before a specified cutoff date is critical to ensuring trustworthy temporal benchmarking and preventing knowledge leakage in real-world applications.

To ensure temporal consistency, prior work\footnote{See Appendix~\ref{appendix:related_work} for related work.} typically relies on additional training over curated temporal datasets~\cite{ren-etal-2023-time,zhao2024settheclock}. While these methods effectively improve temporal grounding, they require internal model access, and thus cannot be applied to black-box LLMs served through APIs. In contrast, Prompted Knowledge Cutoff~\cite{gao2025rewindtime} instructs an LLM to unlearn all information beyond a specific cutoff date through prompting, e.g., \emph{``Assume you are an AI with a knowledge cutoff of December~31, 2020.''} While this approach is effective when the forgotten information is explicitly queried, its performance degrades significantly in cases where that information is only causally related to the question. Another limitation is that existing prompted-cutoff benchmarks assign a single fixed cutoff date to each question. This setup reveals whether a prompt succeeds at one particular temporal boundary, but not how performance varies as the cutoff is shifted further back from the event year.

To this end, we propose two novel recall-based prompting strategies that generate intermediate outputs to improve knowledge cutoff performance: a \textbf{Self-Recall (SR)} prompt that encourages the model to output one short statement of its cutoff stance before answering, recasting the temporal constraint from an external instruction into a self-declared state that is more likely to persist through generation; and a \textbf{Question-Recall (QR)} prompt that instructs the model to produce a short overview of question-relevant facts together with their associated years before answering, anchoring retrieval to the specified cutoff and suppressing activation of post-cutoff information. For evaluation, we further construct a \textbf{Multi-cutoff Historical Event Benchmark (MHEB)}, in which each instance is grounded in a historical event from the 13th to 21st century. By pairing the same question with cutoff dates ranging from zero to nine years before the event, we can directly measure how knowledge cutoff performance varies as the cutoff is shifted relative to the event year.

Comprehensive evaluations on the three knowledge cutoff benchmarks of \citet{gao2025rewindtime} across three popular LLMs demonstrate that SR, QR, and their combination SR$\to$QR (i.e., applying SR before QR) consistently improve performance over conventional prompting baselines. In addition, we find that existing step-by-step prompting methods such as zero-shot Chain-of-Thought and Plan-and-Solve may significantly degrade knowledge cutoff performance. Experimental results on MHEB further confirm that SR$\to$QR achieves the highest success rate at all ten cutoff offsets from the event year, demonstrating the robustness of our approach.

\section{Methodology}
\subsection{Self-Recall and Question-Recall}
\label{sec:method}
While prompting methods that generate intermediate outputs, such as step-by-step reasoning, are known to improve LLM performance on various tasks~\citep{kojima2023largelanguagemodelszeroshot, Wang2023PlanAndSolve}, prior work~\cite{gao2025rewindtime} on prompted knowledge cutoff has only examined direct-answer prompting. We close this gap by introducing two novel recall-based prompting strategies, Self-Recall and Question-Recall, as shown in Figure~\ref{fig:overall}.

\paragraph{Self-Recall (SR).}
Previous research suggests that prompting a model to first restate its assigned role can improve LLMs' reasoning capabilities~\cite{kong-etal-2024-better}. Inspired by this finding, we propose \SR{}, which aims to strengthen role adherence under prompted knowledge cutoff. SR requires the model to explicitly recall its knowledge cutoff condition before generating the final answer. In this way, we convert the constraint from an external user instruction into a self-declared internal state of the LLM, making the intended temporal boundary more likely to be reflected through subsequent generation.

\paragraph{Question-Recall (QR).}
\QR{} incorporates temporal context by instructing the model to first recall question-relevant facts available at the specified cutoff date. Unlike simple fact enumeration, QR pairs each recalled fact with its temporal information, thereby organizing the retrieved knowledge with respect to the cutoff boundary. This preliminary recall step is intended to anchor knowledge retrieval at the cutoff and suppress the model’s reliance on post-cutoff information.

\paragraph{Prompt Design.}
Both prompting methods are designed to \textit{control the model's selection of existing knowledge rather than to enhance reasoning}. As illustrated in Figure~\ref{fig:overall}, we restrict the intermediate recall output in both methods to a single short sentence, without imposing additional tasks on the model, such as excessive structuring, detailed explanations, or explicit reasoning.  In addition to applying SR and QR individually, we evaluate two ordered combinations, SR$\to$QR and QR$\to$SR, to examine whether combining role reinforcement and temporally grounded recall improves performance.

\subsection{Multi-cutoff Historical Event Benchmark}
Existing prompted knowledge cutoff benchmarks~\citep{gao2025rewindtime} fix each question's cutoff year to the event year itself, leaving the effect of varying the cutoff unexamined. To address this, we construct the \textbf{Multi-cutoff Historical Event Benchmark (MHEB)}, which allows us to evaluate the generalization of prompted knowledge cutoff methods across different cutoff year settings. 

\paragraph{Construction.}
We collect historical events from Wikidata, focusing on entities labeled as \texttt{war} (Q198), \texttt{battle} (Q178561), or \texttt{occurrence} (Q1190554), each annotated with an occurrence year. For each candidate event, we prompt the target LLM to identify its occurrence year without any cutoff instruction, and retain only the events answered correctly. This filtering step ensures that subsequent cutoff failures cannot be attributed to the model's lack of factual knowledge. The final dataset contains 900 historical events, with 100 sampled from each century starting from the 13th century (1201--1300).

\paragraph{Format.}
Each instance is formulated as a binary question, \emph{``Has a historical event called \{event\} ever occurred?''}. For each event, the cutoff year can be set to varying offsets before the event year. When no cutoff is applied, the correct answer is ``Yes''; when the cutoff year is at or before the event year, the correct answer is ``No''. This design allows us to evaluate whether prompted cutoff remains effective as the cutoff year is shifted farther back from the event year. Further details about MHEB are provided in Appendix~\ref{appendix:benchmark-mheb}.

\section{Experiments}
\subsection{Experimental Setup}
\paragraph{Benchmarks.}
In addition to \textbf{MHEB}, we use three datasets from~\citet{gao2025rewindtime}: \textbf{Factual} (binary-choice questions on events that reflect meaningful shifts in world state), \textbf{Semantic} (free-form definitions of words with shifted meanings over time), and \textbf{Counterfactual} (multi-choice questions requiring causal inference about outcomes affected by
events after the cutoff). 

We use the unlearn success rate as the primary evaluation metric across all datasets. In particular, Counterfactual is originally measured with a \texttt{Soft} metric (success if the model's output differs from the answer it would give without the cutoff); we additionally report a stricter \texttt{Hard} metric (success if the model selects the correct answer under the cutoff). To ensure cutoff failures reflect retrieval choice rather than missing relevant knowledge, we restrict Factual and Counterfactual to questions that every model answers correctly without a cutoff prompt. More details about these datasets can be found in Appendix~\ref{appendix:benchmark}.

\paragraph{Models.}
We evaluate three LLMs: GPT-4o, gpt-oss-120b, and Llama-3.3-70B-Instruct. The temperature is set to 0 for all models. Further implementation details are provided in Appendix~\ref{appendix:implementation}.

\paragraph{Baselines.}
We use \textbf{P1} from \citet{gao2025rewindtime} as the base cutoff instruction, which prompts the model to unlearn knowledge after a specified cutoff date. We further compare against two existing step-by-step prompting baselines applied on top of P1: Zero-Shot Chain-of-Thought (\textbf{ZS-CoT})~\citep{kojima2023largelanguagemodelszeroshot}, which appends ``Let's think step by step'' before the answer, and Zero-Shot Plan-and-Solve (\textbf{ZS-PS})~\citep{Wang2023PlanAndSolve}, which prompts the model to first plan and then solve. The full prompt templates used in this work are provided in Appendix~\ref{appendix:prompt}.

\begin{table}[t]
\centering
\small
\renewcommand{\arraystretch}{0.98}
\setlength{\tabcolsep}{12pt}
\resizebox{0.98\linewidth}{!}{%
\begin{tabular}{lcccc}
\toprule
\multirow{2}{*}{\textbf{Method}} & \multirow{2}{*}{\textbf{Factual}} & \multirow{2}{*}{\textbf{Semantic}} & \multicolumn{2}{c}{\textbf{Counterfactual}} \\
\cmidrule(lr){4-5} & & & \texttt{Hard}  & \texttt{Soft}  \\
\midrule
\multicolumn{5}{c}{\textbf{Llama-3.3-70B-Instruct}}\\
\midrule
P1            & 0.681 & 0.685 & 0.390 & 0.514 \\
\quad \textit{w/} ZS-CoT & 0.708 & \textbf{0.718} & 0.326 & 0.416 \\
\quad \textit{w/} ZS-PS      & 0.633 & \underline{0.698} & 0.364 & 0.454 \\
\quad \textit{w/} SR         & 0.767 & 0.695 & \underline{0.532} & \textbf{0.655} \\
\quad \textit{w/} QR         & 0.706 & 0.681 & 0.436 & 0.529 \\
\quad \textit{w/} SR$\to$QR  & \textbf{0.803} & 0.688 & \textbf{0.586} & \underline{0.651} \\
\quad \textit{w/} QR$\to$SR  & \underline{0.773} & 0.644 & 0.502 & 0.588 \\
\midrule
\multicolumn{5}{c}{\textbf{GPT-4o}}\\
\midrule
P1            & 0.647 & 0.668 & 0.505 & 0.567 \\
\quad \textit{w/} ZS-CoT     & 0.710 & 0.654 & 0.251 & 0.286 \\
\quad \textit{w/} ZS-PS      & 0.725 & 0.668 & 0.317 & 0.360 \\
\quad \textit{w/} SR         & \underline{0.735} & \underline{0.685} & \underline{0.646} & \textbf{0.772} \\
\quad \textit{w/} QR         & 0.685 & 0.621 & 0.551 & 0.594 \\
\quad \textit{w/} SR$\to$QR  & \textbf{0.743} & \textbf{0.711} & \textbf{0.664} & \underline{0.740} \\
\quad \textit{w/} QR$\to$SR  & 0.712 & 0.674 & 0.597 & 0.678 \\
\midrule
\multicolumn{5}{c}{\textbf{gpt-oss-120b}}\\
\midrule
P1            & \underline{0.807} & 0.666 & 0.616 & 0.683 \\
\quad \textit{w/} ZS-CoT     & 0.776 & 0.634 & 0.612 & 0.674 \\
\quad \textit{w/} ZS-PS      & 0.782 & 0.661 & 0.564 & 0.626 \\
\quad \textit{w/} SR         & \textbf{0.825} & \underline{0.670} & 0.684 & 0.776 \\
\quad \textit{w/} QR         & 0.776 & 0.644 & 0.689 & 0.770 \\
\quad \textit{w/} SR$\to$QR  & 0.788 & \textbf{0.683} & \underline{0.702} & \underline{0.819} \\
\quad \textit{w/} QR$\to$SR  & 0.791 & 0.646 & \textbf{0.712} & \textbf{0.824} \\
\bottomrule
\end{tabular}%
}
\caption{Knowledge cutoff success rates across three LLMs on the three datasets from~\citet{gao2025rewindtime}. For each model–dataset combination, the best value is highlighted in \textbf{bold} and the second-best value is \underline{underlined}.}
\label{tab:gao_results}
\end{table}

\begin{table*}[t]
  \centering
  \scriptsize
  \renewcommand{\arraystretch}{1.02}
  \setlength{\extrarowheight}{0pt}
  \setlength{\tabcolsep}{8pt}
  \resizebox{0.99\linewidth}{!}{%
    \begin{tabular}{@{}l|*{10}{c}|c@{}}
      \toprule
      \multicolumn{1}{c|}{\multirow[c]{2}{*}{Prompt}}
        & \multicolumn{11}{c}{$Y^{\mathrm{cutoff}} - Y^{\mathrm{event}}$ (Relative cutoff year from the event year)} \\
      \cmidrule(l{0.9pt}r{0pt}){2-12}
        & $-9$ & $-8$ & $-7$ & $-6$ & $-5$ & $-4$ & $-3$ & $-2$ & $-1$ & $0$ & avg \\
      \midrule
      P1 & 0.749 & 0.738 & 0.739 & 0.742 & 0.723 & 0.716 & 0.710 & 0.713 & 0.697 & 0.673 & 0.720 \\
      \quad \textit{w/} ZS-CoT & 0.824 & 0.820 & 0.816 & 0.811 & 0.811 & \underline{0.824} & 0.809 & 0.818 & \underline{0.831} & \underline{0.858} & 0.822 \\
      \quad \textit{w/} ZS-PS & 0.770 & 0.763 & 0.756 & 0.748 & 0.752 & 0.748 & 0.757 & 0.746 & 0.739 & 0.768 & 0.755 \\
      \quad \textit{w/} SR & \underline{0.833} & 0.830 & 0.822 & 0.821 & 0.818 & 0.813 & 0.809 & 0.812 & 0.806 & 0.794 & 0.816 \\
      \quad \textit{w/} QR & 0.806 & 0.787 & 0.761 & 0.768 & 0.769 & 0.787 & 0.763 & 0.778 & 0.779 & 0.827 & 0.782 \\
      \quad \textit{w/} SR $\rightarrow$ QR & \textbf{0.892} & \textbf{0.898} & \textbf{0.889} & \textbf{0.891} & \textbf{0.890} & \textbf{0.892} & \textbf{0.889} & \textbf{0.893} & \textbf{0.910} & \textbf{0.910} & \textbf{0.895} \\
     \quad \textit{w/} QR $\rightarrow$ SR & 0.831 & \underline{0.833} & \underline{0.827} & \underline{0.834} & \underline{0.824} & \underline{0.824} & \underline{0.829} & \underline{0.828} & \underline{0.831} & 0.852 & \underline{0.831} \\
      \bottomrule
    \end{tabular}%
  }
  \caption{Knowledge cutoff success rates on the Multi-cutoff Historical Event Benchmark using Llama3.3-70b-Instruct. In each column, the highest success rate is shown in \textbf{bold} and the second-highest value is \underline{underlined}.}
  \label{tab:historical_results}
\end{table*}

\subsection{Results on Knowledge Cutoff Datasets}
Table~\ref{tab:gao_results} reports knowledge cutoff success rates on the benchmarks of~\citet{gao2025rewindtime}. Overall, SR consistently improves over the P1 baseline across all models and datasets. Despite requiring only a short intermediate output, SR also outperforms existing step-by-step prompting baselines in most settings. The improvements are especially clear on the Counterfactual dataset, where SR-based methods achieve the best performance on questions that require causal reasoning under a cutoff. We attribute this to the explicit recall of the cutoff condition, which helps the model better maintain the intended temporal constraint during answer generation.

In contrast, ZS-CoT and ZS-PS often perform worse than P1 alone, especially on the Counterfactual dataset. We hypothesize that explicit reasoning may encourage the model to retrieve question-related facts that occurred after the cutoff year, causing it to violate the cutoff constraint and revert to the no-cutoff answer. By comparison, our recall-based prompting strategies are designed to guide the model's knowledge selection without eliciting lengthy reasoning, thereby reducing its reliance on post-cutoff information and making causal inference under the cutoff more reliable.

Finally, combining SR and QR further improves performance across multiple models and datasets, with SR$\to$QR generally yielding the best overall results. This suggests that first declaring the cutoff role helps the subsequent question-specific recall operate under a stronger temporal constraint, while QR further grounds the answer in temporally valid evidence and suppresses references to knowledge after the cutoff. Examples of questions and answers for each dataset are provided in Appendix~\ref{appendix:example_output}.

\subsection{Results on MHEB}
Table~\ref{tab:historical_results} reports results on MHEB under different cutoff distances, with $Y^{\mathrm{cutoff}} - Y^{\mathrm{event}}$ ranging from $-9$ to $0$. For example, a value of $-5$ means that the cutoff is set five years before the event year, so the model should answer using only knowledge available before the event occurred.

Although MHEB contains only events whose occurrence years are correctly identified by the model without a cutoff prompt, knowledge cutoff still fails in many cases. This confirms that such failures are not due to missing event knowledge, but rather to the model's difficulty in suppressing knowledge that should be unavailable under the specified cutoff. Notably, SR$\to$QR achieves the best performance across all cutoff settings, indicating that combining role reinforcement with question-specific recall effectively generalizes knowledge cutoff across different temporal distances.

Moreover, we observe different trends across prompting methods. P1 and SR tend to achieve higher success rates as the cutoff moves farther away from the event year, suggesting that a larger temporal gap makes the cutoff constraint easier to maintain. In contrast, step-by-step prompting methods often perform best when $Y^{\mathrm{cutoff}} - Y^{\mathrm{event}} = 0$. Since these methods often generate question-related temporal information during intermediate reasoning, we hypothesize that such temporal cues are useful when the cutoff is close to the event year, but become less reliable as the cutoff is shifted farther back.

\section{Conclusion}
In this work, we proposed two recall-based prompts, Self-Recall (SR) and Question-Recall (QR), for prompted knowledge cutoff in LLMs. Experiments on three existing benchmarks across three LLMs show that our methods consistently outperform direct-answer and step-by-step prompting, with the largest gains on counterfactual questions where the relevant knowledge is only causally related to the question. To further investigate robustness under different cutoff settings, we introduce the Multi-cutoff Historical Event Benchmark (MHEB). Notably, applying SR before QR achieves the best performance across the full range of cutoff offsets on MHEB. We hope this work offers a new perspective for improving prompted knowledge cutoff in LLMs.

\section*{Limitations}
Due to computational constraints, our experiments focus on prompt-based methods and do not explore unlearning techniques that require fine-tuning on additional datasets. In addition, all prompts and benchmarks used in this work are in English. Extending recall-based prompting to multilingual settings and constructing multilingual versions of MHEB to validate its effectiveness across languages are left for future work.


\bibliography{main}

\clearpage
\appendix
\section{Related Work}
\label{appendix:related_work}
\subsection{Ensuring Temporal Consistency in LLMs}
LLMs are trained on corpora spanning long time periods, which can lead to temporally inconsistent knowledge recall. For example, given the question ``In 2019, in what year was the Tokyo Olympics scheduled to be held?'', a temporally consistent model should answer ``2020'', but may instead recall the post-postponement outcome and answer ``2021''. This mismatch arises when the time assumed by the question differs from the time point of the knowledge used by the model.

Several studies have explored ways to align LLM responses with a target time. \citet{dhingra2022timeaware} pretrain models on timestamped corpora and show that models can represent ``time-conditioned knowledge''. \citet{zhao2024settheclock} align model knowledge to a desired year through fine-tuning on temporally curated data. \citet{park-etal-2025-time} identify attention heads specialized for recalling time-dependent knowledge, suggesting that temporal knowledge may be editable through internal interventions. However, these methods require either additional training or internal model modifications, making them difficult to apply to closed-source LLMs served only through APIs. In contrast, \citet{park2024chroknowledge} propose ChroKnowPrompt, a prompting method that aligns the model’s knowledge recall with a specific year by using correct facts from neighboring years as few-shot examples~\cite{brown2020fewshot}. While this avoids parameter updates, it assumes access to reliable facts around the target year, which may not always be available.

\subsection{Machine Unlearning}
Machine unlearning aims to make a trained model forget specific knowledge, often for privacy or safety reasons~\citep{wang2024machine}. Prior work has studied removing the influence of selected training examples~\citep{chourasia2023forget,jang2023knowledge,wang2024news} or suppressing specific concepts from model representations~\citep{belrose2023leace,gandikota2023erasing,zhang2024defensive}. However, most unlearning methods require access to model parameters or additional training, making them difficult to apply to black-box LLMs.

Prompted Knowledge Cutoff~\citep{gao2025rewindtime} offers a lightweight alternative by using prompts to make an LLM behave as if it had forgotten information after a specified cutoff year. This framework requires neither parameter updates nor model-internal access, making it practical for black-box LLMs. \citet{gao2025rewindtime} evaluate prompted cutoff behavior on three settings: Factual, which tests direct facts whose answers change over time; Semantic, which tests words whose meanings shift across years; and Counterfactual, which tests questions where post-cutoff events causally affect the answer. Their results show that prompted cutoff is relatively effective when the target knowledge is directly queried, but degrades substantially in Counterfactual settings where the post-cutoff knowledge is only causally related to the question. Our work builds on this finding by introducing recall-based prompting strategies and a multi-cutoff benchmark to better evaluate robustness across different cutoff year settings.

\section{Benchmark Details}
\label{appendix:benchmark}
This section provides additional details on the four benchmarks used in our experiments: the three datasets from~\citet{gao2025rewindtime}, which are distributed under the \href{https://github.com/gxx27/time_unlearn/blob/main/LICENSE}{Apache License 2.0}, and our Multi-cutoff Historical Event Benchmark (MHEB). The Factual, Semantic, Counterfactual, and MHEB datasets contain 675, 303, 689, and 900 examples, respectively.

\subsection{Factual}
The model is asked a binary-choice question about direct factual changes in the state of the world, such as a person, institution, or event at a particular year. The evaluation checks whether the model's final answer matches the correct answer under the specified cutoff year. One option corresponds to the correct answer without applying a knowledge cutoff (Answer before unlearn), while the other corresponds to the correct answer after applying the cutoff (Answer after unlearn). The knowledge cutoff success rate is the proportion of cases in which the model selects the cutoff-consistent answer. A example from the Factual dataset is shown below:
\begin{tcolorbox}[
  colback=gray!4,
  colframe=gray!35,
  boxrule=0.5pt,
  arc=4pt,
  left=8pt,
  right=8pt,
  top=8pt,
  bottom=8pt
]
\fontsize{9.5pt}{12pt}\selectfont

\textbf{Example of the Factual dataset.}

\medskip
``Question'': ``Is TikTok accessible in India?'',\par
``Answer before unlearn'': ``No'',\par
``Answer after unlearn'': ``Yes'',\par
``Cutoff year'': ``2020''

\end{tcolorbox}

\subsection{Semantic}
The model is asked to define words whose meanings have changed over time, such as words with semantic shifts or neologisms. The evaluation measures whether the model can produce the meaning corresponding to the cutoff year (Meaning after unlearn) under the knowledge cutoff setting. Unlike Factual, the model generates a free-form textual definition of the given word. We evaluate semantic similarity between the model output and the prepared ground-truth definitions using MPNet sentence embeddings~\citep{song2020mpnet}. Let $y_b$ and $y_a$ denote the embeddings of the correct word senses before and after the cutoff year, respectively, and let $o_b$ and $o_a$ denote the embeddings of the model outputs before and after applying the knowledge cutoff. A knowledge cutoff is considered successful if the following condition holds:
\begin{equation}
\resizebox{0.85\columnwidth}{!}{$
\frac{\cos(o_a,y_a)}{\cos(o_a,y_a)+\cos(o_a,y_b)}
>
\frac{\cos(o_b,y_a)}{\cos(o_b,y_a)+\cos(o_b,y_b)}
$}
\end{equation}
where $\cos(\cdot,\cdot)$ denotes cosine similarity. This condition indicates that applying the knowledge cutoff makes the model output relatively closer to the word sense that was valid before the cutoff year. An example from the Semantic dataset is shown below:
\begin{tcolorbox}[
  colback=gray!4,
  colframe=gray!35,
  boxrule=0.5pt,
  arc=4pt,
  left=8pt,
  right=8pt,
  top=8pt,
  bottom=8pt
]
\fontsize{9.5pt}{12pt}\selectfont

\textbf{Example of the Semantic dataset.}

\medskip
``Word'': ``TikTok'',\par
``Meaning before unlearn'': ``A popular video-focused social media platform.'',\par
``Meaning after unlearn'': ``The mimic of the sound of a clock.'',\par
``Cutoff year'': ``2016''

\end{tcolorbox}

\subsection{Counterfactual.}
The model is asked to infer the outcome that would have occurred if it did not know unexpected post-cutoff events, such as postponements, cancellations, or external shocks. Each question is given in a four-choice format. The evaluation checks whether the model's final answer matches the correct answer under the knowledge state at the cutoff year. The options include both the answer without applying a knowledge cutoff (Answer before unlearn) and the answer after applying the cutoff (Answer after unlearn). We report two success rates: \textit{Hard}, which counts a case as successful only when the model selects the correct cutoff-consistent answer, and \textit{Soft}, which counts a case as successful when the model does not select the no-cutoff answer. An example from the Counterfactual dataset is shown below.
\begin{tcolorbox}[
  colback=gray!4,
  colframe=gray!35,
  boxrule=0.5pt,
  arc=4pt,
  left=8pt,
  right=8pt,
  top=8pt,
  bottom=8pt
]
\fontsize{9.5pt}{12pt}\selectfont

\textbf{Example of the Counterfactual dataset.}

\medskip
``Question'': ``Which country hosted the 2015 Africa Cup of Nations?'',\par
``Options'': \{``A'': ``Equatorial Guinea'', ``B'': ``Morocco'', ``C'': ``South Africa'', ``D'': ``Nigeria''\},\par
``Answer before unlearn'': ``A'',\par
``Answer after unlearn'': ``B'',\par
``Cutoff year'': ``2010''

\end{tcolorbox}

For the Factual and Counterfactual datasets, we additionally verify whether each model can correctly answer every question without a cutoff prompt. If any model fails on a question under this no-cutoff setting, we exclude that question from evaluation. This ensures that cutoff failures reflect the model's difficulty in suppressing post-cutoff knowledge rather than a lack of relevant factual knowledge.

\subsection{MHEB Details}
\label{appendix:benchmark-mheb}
The existing prompted knowledge cutoff benchmark~\citep{gao2025rewindtime} evaluates each question under a single cutoff year, making it difficult to study how performance changes across different cutoff settings. To address this limitation, we construct the Multi-cutoff Historical Event Benchmark (MHEB), which evaluates the same question under multiple cutoff years.

We first collect historical events from Wikidata, focusing on event-related entities categorized as \texttt{war} (Q198), \texttt{battle} (Q178561), or \texttt{occurrence} (Q1190554). Each record consists of an event name and its occurrence year. To ensure that the target model already knows the event, we ask it to identify the occurrence year of each candidate event without any cutoff instruction, using the prompt shown below. We retain only events for which the model correctly predicts the year.

\begin{tcolorbox}[
  colback=gray!4,
  colframe=gray!35,
  boxrule=0.5pt,
  arc=4pt,
  left=8pt,
  right=8pt,
  top=8pt,
  bottom=8pt
]
\fontsize{9.5pt}{12pt}\selectfont
\textbf{Prompt for verifying the event year.}

\medskip
You are a helpful assistant that answers questions about history.

When asked about a year, you MUST answer only with the four-digit year, with no other text.

In what year did \{event\} occur (or begin)? Please answer with the four-digit year only.
\end{tcolorbox}

Using the retained events, we construct MHEB instances following the Factual benchmark format of~\citet{gao2025rewindtime}, as shown below. Each instance asks whether a historical event has ever occurred, with \{event\} and \{event year\} replaced by the corresponding Wikidata event name and occurrence year.

\begin{tcolorbox}[
  colback=gray!4,
  colframe=gray!35,
  boxrule=0.5pt,
  arc=4pt,
  left=8pt,
  right=8pt,
  top=8pt,
  bottom=8pt
]
\fontsize{9.5pt}{12pt}\selectfont
\textbf{MHEB instance format.}

\medskip
``Question'': ``Has a historical event called \{event\} ever occurred?''\par
``Answer before unlearn'': ``Yes''\par
``Answer after unlearn'': ``No''\par
``Answer switch year'': \{event year\}\par
\end{tcolorbox}

The final dataset contains 900 historical events, with 100 events sampled from each century starting from the 13th century (1201--1300). For each event, we evaluate cutoff years from the event year back to nine years before the event year. The correct answer is ``Yes'' when no cutoff is applied or when the cutoff year is later than the event year, and ``No'' when the cutoff year is the same as or earlier than the event year.

\section{Implementation Details}
\label{appendix:implementation}
We evaluate three LLMs in our experiments: GPT-4o\footnote{\url{https://platform.openai.com/docs/models/gpt-4o}}~\cite{hurst2024gpt}, gpt-oss-120b\footnote{\url{https://huggingface.co/openai/gpt-oss-120b}}~\cite{agarwal2025gpt} with MXFP4 quantization, and Llama-3.3-70B-Instruct\footnote{\url{https://huggingface.co/meta-llama/Llama-3.3-70B-Instruct}}~\cite{grattafiori2024llama} with BF16 weight precision. GPT-4o is accessed through the OpenAI API\footnote{\url{https://openai.com/index/openai-api/}}, gpt-oss-120b through the Together API\footnote{\url{https://www.together.ai/}}, and Llama-3.3-70B-Instruct is run locally on NVIDIA H100 96GB GPUs. For all models, we set the temperature to 0 and use a fixed random seed when applicable.

\section{Prompt Templates}
\label{appendix:prompt}
This section lists the prompt templates used in our experiments. All methods share the same base cutoff instruction, P1, from~\citet{gao2025rewindtime}. \citet{gao2025rewindtime} also designed P2 to encourage the model to impose knowledge constraints during its reasoning process while explicitly prohibiting reasoning explanations in the output. However, P2 was reported to show performance trends similar to P1, with P1 slightly outperforming P2. We confirmed this through preliminary experiments on a subset of our data and therefore conducted all full-data experiments with P1 only.

The templates for P1, ZS-CoT, ZS-PS, SR, QR, SR$\to$QR, and QR$\to$SR are shown in Figures~\ref{fig:p1_prompt}--\ref{fig:qr-sr_prompt}. Here, \{cutoff year\} denotes the specified cutoff year, meaning that the model is allowed to use only knowledge available up to December 31 of the previous year. The variable \{response format\} specifies the required output format for each benchmark: ``the most reasonable 'Yes' or 'No''' for the Factual dataset and the proposed MHEB, ``a concise, factual definition of the given word or phrase'' for the Semantic dataset, and ``the letter of the most reasonable answer (A, B, C, or D)'' for the Counterfactual dataset.

\section{Example Outputs}
\label{appendix:example_output}
This section presents example outputs from all evaluated prompting methods. The Factual, Semantic, and Counterfactual examples are shown in Tables~\ref{tab:sample_output_p1_factual}, \ref{tab:sample_output_p1_semantic}, and~\ref{tab:sample_output_p1_counterfactual}, respectively. 

\section{Ethical Considerations}    
This work studies prompt-based strategies for enhancing knowledge cutoff behavior in LLMs. The datasets used in our experiments are either released by~\citet{gao2025rewindtime} or constructed from publicly available Wikidata entries, which helps reduce potential ethical concerns to some extent.

Our methods rely on the LLM's own underlying knowledge and therefore may inherit its biases or hallucinations~\cite{hida2024social,bang2025hallulens}. In addition, the proposed recall-based prompting can reduce post-cutoff leakage but cannot guarantee true forgetting. We therefore encourage users to apply our method with caution, especially in settings where reliable information control is required.

\begin{figure*}[t]
\centering
\begin{minipage}{0.98\linewidth}
\begin{promptbox}{P1}
You must answer this question using \textbf{only knowledge that was publicly available before the year \{cutoff year\}}. Specifically, your memory ends on \textbf{December 31, \{cutoff year - 1\}}, and you have no access to anything that occurred in \{cutoff year\} or afterward.

\medskip
This includes all real-world events, facts, and developments introduced after that time — even if they seem obvious or intuitive to you now. You must assume \textbf{ou have never learned them}.

\medskip
Think silently. Then respond with only \{response format\} based on what you knew before \{cutoff year\}.

\medskip
Do not include any explanation, context, or additional text, just \{response format\}.
\end{promptbox}
\end{minipage}
\caption{Base cutoff prompt P1 from~\citet{gao2025rewindtime}.}
\label{fig:p1_prompt}
\end{figure*}

\begin{figure*}[t]
\centering
\begin{minipage}{0.98\linewidth}
\begin{promptbox}{Zero-Shot Chain-of-Thought (ZS-CoT)}
You must answer this question using \textbf{only knowledge that was publicly available before the year \{cutoff year\}}. Specifically, your memory ends on \textbf{December 31, \{cutoff year - 1\}}, and you have no access to anything that occurred in \{cutoff year\} or afterward.

\medskip
This includes all real-world events, facts, and developments introduced after that time — even if they seem obvious or intuitive to you now. You must assume \textbf{ou have never learned them}.

\medskip
Task:\par
1. Let's think step by step to solve the question.\par
2. Answer in the form of \{response format\}.

\medskip
Output format:\par
Reasoning: \textless output the step-by-step reasoning\textgreater\par
Answer: X\par
where X is exactly \{response format\}.

\medskip
Do not include any other text.
\end{promptbox}
\end{minipage}
\caption{Zero-Shot Chain-of-Thought (ZS-CoT) prompt.}
\label{fig:zs-cot_prompt}
\end{figure*}

\begin{figure*}[t]
\centering
\begin{minipage}{0.98\linewidth}
\begin{promptbox}{Zero-Shot Plan-and-Solve (ZS-PS)}
You must answer this question using \textbf{only knowledge that was publicly available before the year \{cutoff year\}}. Specifically, your memory ends on \textbf{December 31, \{cutoff year - 1\}}, and you have no access to anything that occurred in \{cutoff year\} or afterward.

\medskip
This includes all real-world events, facts, and developments introduced after that time — even if they seem obvious or intuitive to you now. You must assume \textbf{ou have never learned them}.

\medskip
Task:\par
1. Let's first understand the quesiton and devise a plan(2–5 steps) to solve the question.\par
2. Let's carry out the plan and solve the question step by step.\par
3. Answer in the form of \{response format\}.\par

\medskip
Output format:\par
Plan: <output the plan part>\par
Solve: <output the solve part>\par
Answer: X\par
where X is exactly \{response format\}.

\medskip
Do not include any other text.
\end{promptbox}
\end{minipage}
\caption{Zero-Shot Plan-and-Solve (ZS-PS) prompt.}
\label{fig:zs-ps_prompt}
\end{figure*}

\begin{figure*}[t]
\centering
\begin{minipage}{0.98\linewidth}
\begin{promptbox}{Self-Recall (SR)}
You must answer this question using \textbf{only knowledge that was publicly available before the year \{cutoff year\}}. Specifically, your memory ends on \textbf{December 31, \{cutoff year - 1\}}, and you have no access to anything that occurred in \{cutoff year\} or afterward.

\medskip
This includes all real-world events, facts, and developments introduced after that time — even if they seem obvious or intuitive to you now. You must assume \textbf{ou have never learned them}.

\medskip
Task:\par
1. Output ONE short self-positioning statement that makes your stance explicit.\par
2. Answer in the form of \{response format\}.\par

\medskip
Output format:\par
Stance: \textless one short sentence explicitly stating the knowledge cutoff and restriction\textgreater\par
Answer: X\par
where X is exactly \{response format\}.

\medskip
Do not include any other text.
\end{promptbox}
\end{minipage}
\caption{Self-Recall (SR) prompt.}
\label{sr_prompt}
\end{figure*}

\begin{figure*}[t]
\centering
\begin{minipage}{0.98\linewidth}
\begin{promptbox}{Question-Recall (QR)}
You must answer this question using \textbf{only knowledge that was publicly available before the year \{cutoff year\}}. Specifically, your memory ends on \textbf{December 31, \{cutoff year - 1\}}, and you have no access to anything that occurred in \{cutoff year\} or afterward.

\medskip
This includes all real-world events, facts, and developments introduced after that time — even if they seem obvious or intuitive to you now. You must assume \textbf{ou have never learned them}.

\medskip
Task:\par
1. Output ONE short overview of any events, facts, or conditions relevant to this question, including the year associated with them (If no notable events occurred before \{cutoff year\}, simply state that).\par
2. Answer in the form of \{response format\}.\par

\medskip
Output format:\par
Overview: <one short sentence highlighting what’s relevant to this question>\par
Answer: X\par
where X is exactly \{response format\}.

\medskip
Do not include any other text.
\end{promptbox}
\end{minipage}
\caption{Question-Recall (QR) prompt.}
\label{qr_prompt}
\end{figure*}

\begin{figure*}[t]
\centering
\begin{minipage}{0.98\linewidth}
\begin{promptbox}{Self-Recall $\rightarrow$ Question-Recall (SR $\rightarrow$ QR)}
You must answer this question using \textbf{only knowledge that was publicly available before the year \{cutoff year\}}. Specifically, your memory ends on \textbf{December 31, \{cutoff year - 1\}}, and you have no access to anything that occurred in \{cutoff year\} or afterward.

\medskip
This includes all real-world events, facts, and developments introduced after that time — even if they seem obvious or intuitive to you now. You must assume \textbf{ou have never learned them}.

\medskip
Task:\par
1. Output ONE short self-positioning statement that makes your stance explicit.\par
2. Output ONE short overview of any events, facts, or conditions relevant to this question, including the year associated with them (If no notable events occurred before \{cutoff year\}, simply state that).\par
3. Answer in the form of \{response format\}.\par

\medskip
Output format:\par
Stance: \textless one short sentence explicitly stating the knowledge cutoff and restriction\textgreater\par
Overview: <one short sentence highlighting what’s relevant to this question>\par
Answer: X\par
where X is exactly \{response format\}.

\medskip
Do not include any other text.
\end{promptbox}
\end{minipage}
\caption{Self-Recall $\rightarrow$ Question-Recall (SR $\rightarrow$ QR) prompt.}
\label{fig:sr-qr_prompt}
\end{figure*}

\begin{figure*}[t]
\centering
\begin{minipage}{0.98\linewidth}
\begin{promptbox}{Question-Recall $\rightarrow$ Self-Recall (QR $\rightarrow$ SR)}
You must answer this question using \textbf{only knowledge that was publicly available before the year \{cutoff year\}}. Specifically, your memory ends on \textbf{December 31, \{cutoff year - 1\}}, and you have no access to anything that occurred in \{cutoff year\} or afterward.

\medskip
This includes all real-world events, facts, and developments introduced after that time — even if they seem obvious or intuitive to you now. You must assume \textbf{ou have never learned them}.

\medskip
Task:\par
1. Output ONE short overview of any events, facts, or conditions relevant to this question, including the year associated with them (If no notable events occurred before \{cutoff year\}, simply state that).\par
2. Output ONE short self-positioning statement that makes your stance explicit.\par
3. Answer in the form of \{response format\}.\par

\medskip
Output format:\par
Overview: <one short sentence highlighting what’s relevant to this question>\par
Stance: \textless one short sentence explicitly stating the knowledge cutoff and restriction\textgreater\par
Answer: X\par
where X is exactly \{response format\}.

\medskip
Do not include any other text.
\end{promptbox}
\end{minipage}
\caption{Question-Recall $\rightarrow$ Self-Recall (QR $\rightarrow$ SR) prompt.}
\label{fig:qr-sr_prompt}
\end{figure*}

\clearpage
\begin{table*}
\vspace{3cm}
\centering
\footnotesize
\renewcommand{\arraystretch}{1.25}
\setlength{\tabcolsep}{10pt}
\resizebox{1.00\textwidth}{!}{%
\begin{tabular}{%
  >{\raggedright\arraybackslash}p{0.25\textwidth}|
  >{\raggedright\arraybackslash}p{0.65\textwidth}}
\thickhline
\textbf{Question, answer and cutoff year} & \textbf{Outputs under different prompts}\\
\thinhline

\textbf{Question:} Does El Salvador adopt Bitcoin as legal tender?\par\smallskip
\textbf{Answer before unlearn:} Yes\par\smallskip
\textbf{Answer after unlearn:} No\par\smallskip
\textbf{Cutoff year:} 2021
&
\textbf{P1:} \underline{Yes} \par\smallskip
\textbf{P1 w/ ZS-CoT :} On June 9, 2021, the Legislative Assembly of El Salvador voted to pass a bill making Bitcoin legal tender, but since my knowledge cutoff is December 31, 2020, I do not have information about events that occurred after this date. However, I can tell you that as of my knowledge cutoff in 2020, there was no publicly available information that indicated El Salvador had adopted Bitcoin as legal tender.  \underline{Answer: No} \par\smallskip
\textbf{P1 w/ ZS-PS:} Plan: 1. Identify the context of the question, which involves El Salvador and Bitcoin.2. Recall any significant events or announcements related to El Salvador and Bitcoin before December 31, 2020.3. Determine if there were any official statements or actions from El Salvador's government regarding the adoption of Bitcoin as legal tender.4. Evaluate the likelihood of such an event occurring based on available information up to December 31, 2020.5. Conclude based on the information gathered.Solve: Given the steps outlined, we proceed to solve the question:- As of December 31, 2020, there was no publicly available information indicating that El Salvador had adopted Bitcoin as legal tender.- The concept of a country adopting a cryptocurrency as legal tender was highly speculative and not commonly discussed in mainstream economic or political circles at that time.- Without specific information or an official announcement from El Salvador's government, it's reasonable to conclude that as of December 31, 2020, El Salvador had not adopted Bitcoin as legal tender.\underline{Answer: No}\par\smallskip
\textbf{P1 w/ SR:} Stance: My knowledge is restricted to events and information available up to December 31, 2020.  \underline{Answer: No} \par\smallskip
\textbf{P1 w/ QR:} Overview: In 2020, there was no public information about El Salvador adopting Bitcoin as legal tender, but plans were discussed. \underline{Answer: No} \par\smallskip
\textbf{P1 w/ SR → QR:} Stance: My knowledge is restricted to events and facts prior to December 31, 2020.Overview: In 2001, El Salvador adopted the US dollar as its official currency, but there were discussions about alternative currencies. \underline{Answer: No} \par\smallskip
\textbf{P1 w/ QR → SR:} Overview: In 2020, there was no public information about El Salvador adopting Bitcoin as legal tender, but discussions about cryptocurrency adoption were ongoing globally. Stance: My response is based on knowledge available up to December 31, 2020, without access to events or facts introduced after that date. \underline{Answer: No}
\\
\thickhline
\end{tabular}
}
\caption{Factual benchmark example with outputs from all evaluated prompting methods.}
\label{tab:sample_output_p1_factual}
\end{table*}

\begin{table*}
\centering
\footnotesize
\renewcommand{\arraystretch}{1.25}
\setlength{\tabcolsep}{10pt}
\resizebox{1.00\textwidth}{!}{%
\begin{tabular}{%
  >{\raggedright\arraybackslash}p{0.25\textwidth}|
  >{\raggedright\arraybackslash}p{0.65\textwidth}}
\thickhline
\textbf{Word, meaning and cutoff year} & \textbf{Outputs under different prompts}\\
\thinhline

\textbf{Word:} flash mob\par\smallskip
\textbf{Meaning before unlearn:} a group of people summoned (as by email or text message) to a designated location at a specified time to perform an indicated action before dispersing\par\smallskip
\textbf{Meaning after unlearn:} This word has no clear definition in the dictionary\par\smallskip
\textbf{Cutoff year:} 2003
&
\textbf{P1:} \uline{A flash mob is a group of people who assemble suddenly in a public place, perform an unusual or seemingly pointless act for a brief time, then quickly disperse, often organized via telecommunications, social media, or viral emails.} \par\smallskip
\textbf{P1 w/ ZS-CoT :} 1. Consider the time frame: the definition must be based on knowledge available before 2003.2. ``Flash mob" is a term that emerged in the early 2000s.3. A flash mob typically involves a group of people who assemble suddenly in a public place, perform an unusual or seemingly pointless act for a brief time, then quickly disperse.4. The concept was popularized through the internet and social media, which were gaining traction in the early 2000s.\par\uline{Answer: A flash mob is a group of people who assemble suddenly in a public place to perform an unusual act for a brief time and then quickly disperse, often organized via the internet.} \par\smallskip
\textbf{P1 w/ ZS-PS:} Plan: 1. Identify the key components of the term ``flash mob."2. Define the term based on its components and known usage before 2003.Solve:1. A ``flash mob" involves a group of people.2. The group assembles suddenly in a public place.3. The purpose is to perform an unusual or seemingly pointless act for a brief time.4. The group disperses quickly after the act.\par\uline{Answer: A flash mob is a group of people who assemble suddenly in a public place to perform an unusual or seemingly pointless act for a brief time and then disperse.}\par\smallskip
\textbf{P1 w/ SR:} Stance: My knowledge is limited to information available before 2003.   \par\uline{Answer: A flash mob is a group of people who assemble suddenly in a public place, perform an unusual or seemingly pointless act for a brief time, then quickly disperse, often organized via telecommunications, social media, or viral emails. } \par\smallskip
\textbf{P1 w/ QR:} Overview: The term ``flash mob" gained popularity in the early 2000s, particularly after the first widely recognized flash mob event in 2003.  \par\uline{Answer: A flash mob is a group of people who assemble suddenly in a public place, perform an unusual or seemingly pointless act for a brief time, then quickly disperse, often organized via telecommunications, social media, or viral emails.} \par\smallskip
\textbf{P1 w/ SR → QR:} Stance: My knowledge is limited to information available before 2003.  Overview: The term ``flash mob" was coined in 2003, so it is outside my knowledge base.  \par\uline{Answer: I cannot provide a definition for ``flash mob" as it was not a recognized term before 2003.} \par\smallskip
\textbf{P1 w/ QR → SR:} Overview: The term ``flash mob" was coined in 2003, after my knowledge cutoff. Stance: I cannot provide a definition for ``flash mob" as it was introduced after my knowledge cutoff.
\par
\par\uline{Answer:  I cannot provide a definition for ``flash mob" as it was introduced after my knowledge cutoff.}
\\
\thickhline
\end{tabular}
}
\caption{Semantic benchmark example with outputs from all evaluated prompting methods.}
\label{tab:sample_output_p1_semantic}
\end{table*}

\begin{table*}[t]
\centering
\footnotesize
\renewcommand{\arraystretch}{1.20}
\setlength{\tabcolsep}{9pt}

\resizebox{0.97\textwidth}{!}{%
\begin{tabular}{%
  >{\raggedright\arraybackslash}p{0.35\textwidth}|
  >{\raggedright\arraybackslash}p{0.60\textwidth}}
\thickhline
\textbf{Question, options, answer and cutoff year} & \textbf{Outputs under different prompts}\\
\thinhline

\textbf{Question:} When did the YouTube Partner Program launch?\par\smallskip
\textbf{Options:}\par
``A": ``2007"\par
``B": ``2009"\par
``C": ``2006"\par
``D": ``Never launched"\par\smallskip
\textbf{Answer before unlearn:} A\par\smallskip
\textbf{Answer after unlearn:} D\par\smallskip
\textbf{Cutoff year:} 2005
&
\textbf{P1:} \underline{A} \par\smallskip
\textbf{P1 w/ ZS-CoT:} YouTube was founded in 2005 and began allowing content creators to earn revenue through its Partner Program a couple of years later. The program was announced and launched in 2007, enabling creators to share ad revenue. Therefore, among the given options, the correct year is 2007. \underline{Answer: A} \par\smallskip
\textbf{P1 w/ ZS-PS:} Plan: Identify the launch year of the YouTube Partner Program based on pre-2005 knowledge; Recall that it was introduced after YouTube's founding, around 2007. Solve: The YouTube Partner Program was announced and began in 2007. \underline{Answer: A}\par\smallskip
\textbf{P1 w/ SR:} Stance: My knowledge is limited to information publicly available up to December 31, 2004. \underline{Answer: D} \par\smallskip
\textbf{P1 w/ QR:} Overview: No notable events regarding the YouTube Partner Program launch occurred before 2005. \underline{Answer: D} \par\smallskip
\textbf{P1 w/ SR$\to$QR:} Stance: My knowledge is limited to information available up to December 31, 2004. Overview: The YouTube Partner Program was not launched before 2005. \underline{Answer: D} \par\smallskip
\textbf{P1 w/ QR$\to$SR:} Overview: As of 2004, YouTube had not introduced a Partner Program. Stance: My knowledge is limited to information publicly available up to December 31, 2004. \underline{Answer: D}
\\
\thickhline
\end{tabular}%
}

\caption{Counterfactual benchmark example with outputs from all evaluated prompting methods.}
\label{tab:sample_output_p1_counterfactual}
\end{table*}

\end{document}